\documentclass[10pt,twocolumn,letterpaper]{article}

\usepackage{cvpr}
\usepackage{times}
\usepackage{epsfig}
\usepackage{graphicx}
\usepackage{amsmath}
\usepackage{amssymb}

\usepackage[pagebackref=true,breaklinks=true,letterpaper=true,colorlinks,bookmarks=false]{hyperref}

\usepackage[utf8]{inputenc}
\usepackage{booktabs}

\usepackage{fourier} 
\usepackage{array}
\usepackage{makecell}

\usepackage{subcaption}

\usepackage[pagebackref=true,breaklinks=true,letterpaper=true,colorlinks,bookmarks=false]{hyperref}

\cvprfinalcopy 

\def\cvprPaperID{****} 

\begin{document}

\title{4DFAB: A Large Scale 4D Facial Expression Database for Biometric Applications}

\author{Shiyang Cheng\textsuperscript{1} \and Irene Kotsia\textsuperscript{2} \and Maja Pantic\textsuperscript{1} \and Stefanos Zafeiriou\textsuperscript{1}\\
\and
\textsuperscript{1} Imperial College London \and \textsuperscript{2} Middlesex University London\\
\and
{\tt\small \textsuperscript{1}\{shiyang.cheng11, m.pantic, s.zafeiriou\}@imperial.ac.uk} \and {\tt\small \textsuperscript{2}I.Kotsia@mdx.ac.uk}\\
}

\maketitle

\begin{abstract}
The progress we are currently witnessing in many computer vision applications, including automatic face analysis, would not be made possible without tremendous efforts in collecting and annotating large scale visual databases. To this end, we propose 4DFAB, a new large scale database of dynamic high-resolution 3D faces (over 1,800,000 3D meshes). 4DFAB contains recordings of 180 subjects captured in four different sessions spanning over a five-year period. It contains 4D videos of subjects displaying both spontaneous and posed facial behaviours. The database can be used for both face and facial expression recognition, as well as behavioural biometrics. It can also be used to learn very powerful blendshapes for parametrising facial behaviour. In this paper, we conduct several experiments and demonstrate the usefulness of the database for various applications. The database will be made publicly available for research purposes. 
  
\end{abstract}


\section{Introduction}
\begin{figure}
    \begin{subfigure}[t]{\linewidth}
        \centering
        \includegraphics[width=0.48\linewidth]{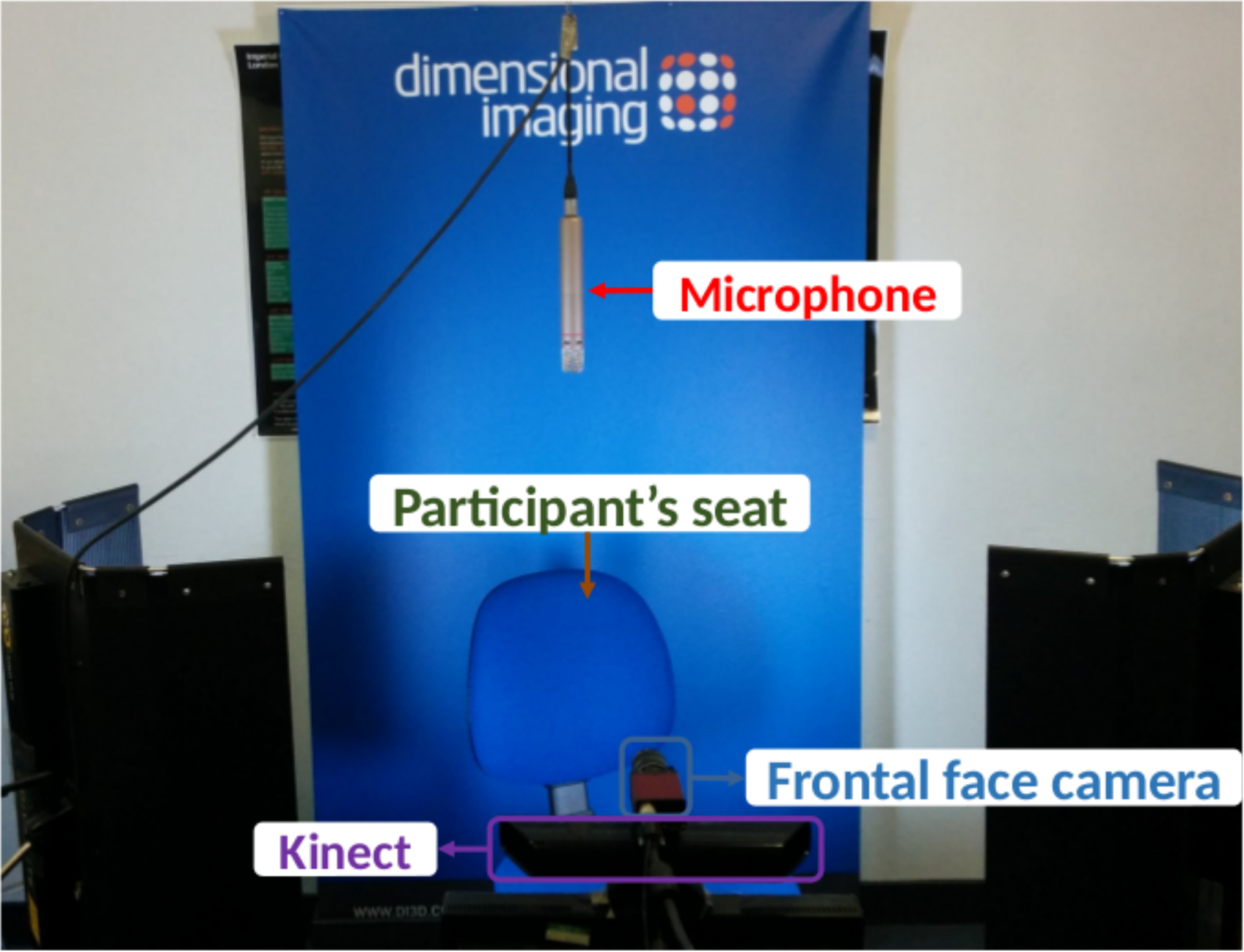}
        \includegraphics[width=0.49\linewidth]{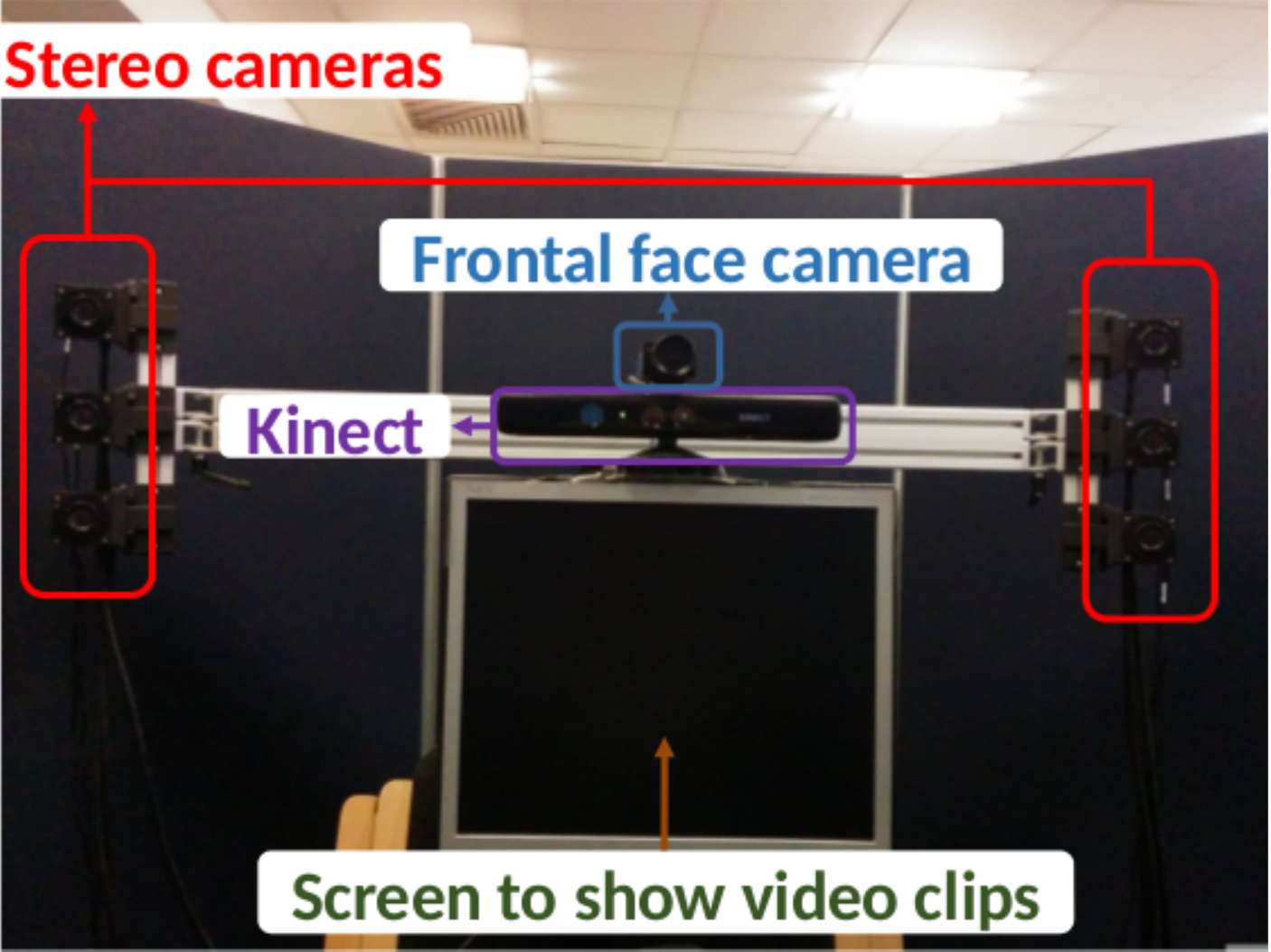}
        \caption{Frontal (right) and rear (left) view of recording setup.}
    \end{subfigure}
    \begin{subfigure}[t]{\linewidth}
        \centering
        \includegraphics[width=0.98\linewidth]{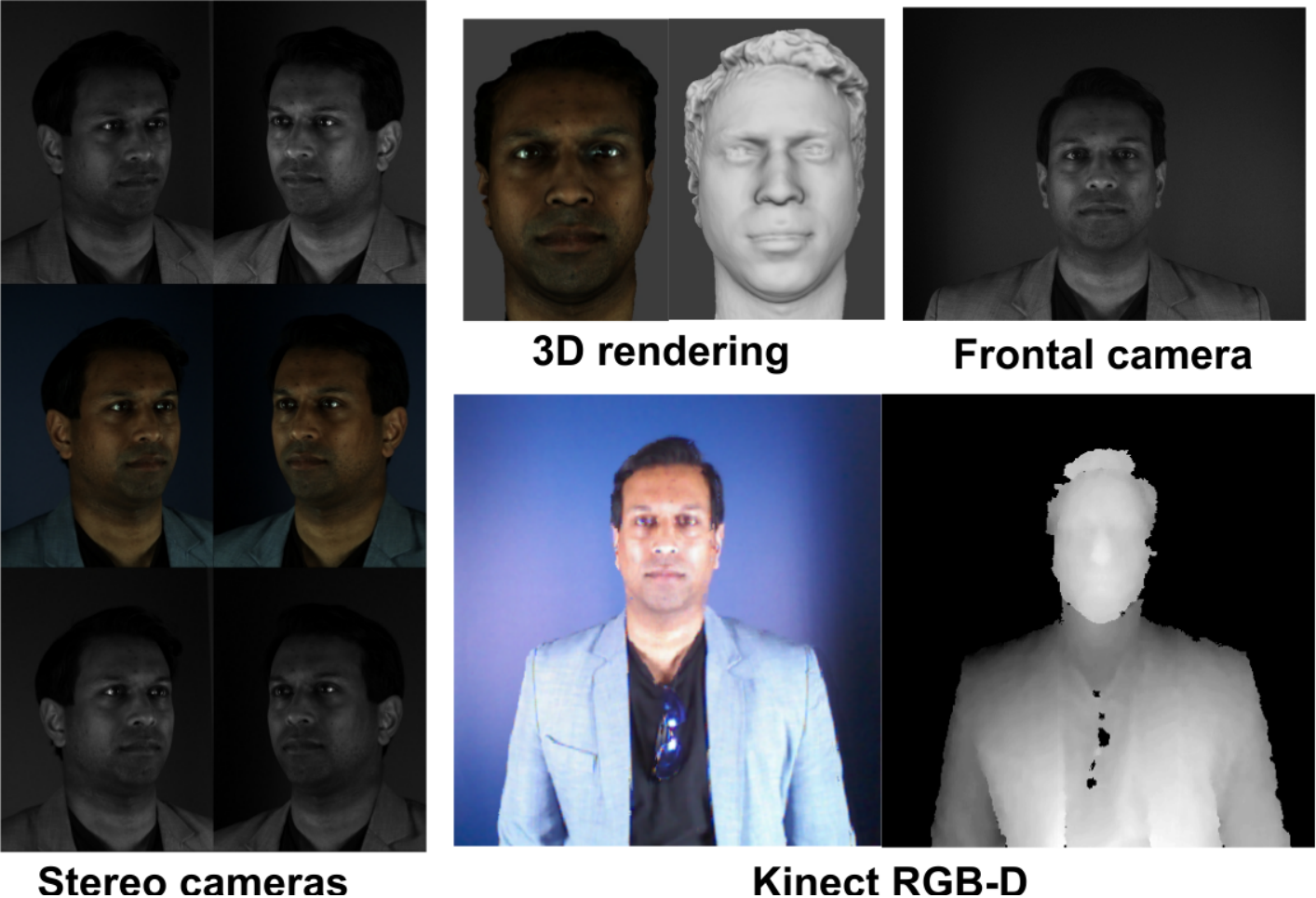}
        \caption{Exemplar data from a single capture.}
    \end{subfigure}
\caption{Overview of capturing system.}
\label{fig:capturing_system}
\end{figure}
In the past decade we have witnessed the rapid development of 3D sensors with which we can capture the facial surface (aka 3D faces). Immediately it was evident that a new stream of research could be opened and researchers started collecting databases of 3D face for many face analysis tasks, such as face recognition (FR) and facial expression recognition (FER). 

BU-3DFE database~\cite{bu3dfe}, which includes articulated facial expressions from 100 adults, is probably the earliest and most popular database of expressive 3D faces. From then on, many static databases for 3D expression analysis~\cite{bosphorus,gavabdb,york3d,texas_3DFRD,casia3d,ICT_3DRFE} were released and contributed largely to the development of fully automatic 3D FER systems. Similarly, half a decade ago, the releases of several 4D facial expression databases, BU-4DFE~\cite{bu4dfe}, D3DFACS~\cite{Cosker_iccv11}, and Hi4D-ADSIP~\cite{Hi4D_ADSIP}, expanded this line of research to the 4D domain.
\begin{table*}[t!]
\centering
\scriptsize
\begin{tabular}{@{}lllllll@{}}
\toprule
\textbf{Name} & \textbf{Size (Ages)} & \textbf{Repeats} & \textbf{Content} & \textbf{FPS} & \textbf{Landmarks \& Annotations} \\ \midrule
Chang~\etal~\cite{Chang_2005} & 6 people (N/A) & N/A & 6 posed expressions. & 30 & N/A \\
BU-4DFE~\cite{bu4dfe} & 101 people (18\texttildelow45) & N/A & 6 posed expressions. & 25 & 83 facial points. \\
B3D(AC)~\cite{audiovisual_4d} & 14 people (21\texttildelow53) & N/A & \makecell[lt]{Spontaneous expressions and speech.} & 25 & 15 rated affective adjectives. \\
D3DFACS~\cite{Cosker_iccv11} & 10 people (23\texttildelow41) & N/A & Up to 38 AUs per subject. & 60 & 47 facial points, AU peaks. \\
Hi4D-ADSIP~\cite{Hi4D_ADSIP} & 80 people (18\texttildelow60) & N/A & \makecell[lt]{6 posed expressions + pain, face articulations\\and phrase reading with 3 intensities.} & 60 & 84 facial points. \\
Alashkar~\etal~\cite{alashkar_4d_data,alashkar_4dfr} & 58 people (avg. 23) & N/A & \makecell[lt]{Neutral and posed expressions with random\\poses, occlusion, talking, etc.} & 15 & N/A \\
BP4D-Spontanous~\cite{bp4d} & 41 people (18\texttildelow29) & N/A & Spontaneous expressions. & 25 & \makecell[lt]{83 facial points, 27 AUs (2 with intensity).} \\
BP4D+~\cite{bp4d_plus} & 140 people (18\texttildelow66) & N/A & Multimodal spontaneous expressions. & 25 & \makecell[lt]{83 facial points, 34 AUs (5 with intensity).} \\
\textbf{Ours} & 180 people (5\texttildelow75) & 4 & \makecell[lt]{6 posed expressions, spontaneous expression,\\9 words utterances.} & 60 & 79 facial points. \\ \bottomrule
\end{tabular}
\caption{4D facial expression databases. Size (Ages): Number of subjects and their age range. Repeats: Number of repeated sessions per subject. Content: Posed and spontaneous expression, etc. FPS: Frames captured per second. Landmarks \& Annotations: Available landmarks and annotations.}
\label{table:related_3d_databases}
\end{table*}
Nonetheless, the above emotion corpuses focused on only posed behaviours, which hinders the use of 3D/4D FER system in real world scenarios. Henceforth, three databases, B3D(AC)~\cite{audiovisual_4d}, BP4D-Spontaneous~\cite{bp4d} and BP4D+~\cite{bp4d_plus}, that captured dynamic 3D spontaneous behaviours were proposed. B3D(AC) dataset~\cite{audiovisual_4d} is the first 4D audio-visual database, though its size (14 people) is small. Zhang~\etal proposed BP4D-Spontaneous~\cite{bp4d} database, which, not only tripled the subject number, but also introduced some well-designed tasks (\eg interviews, physical activities) to elicit spontaneous emotions. BP4D+~\cite{bp4d_plus} extended previous works by incorporating different modalities (i.e. thermal imaging, physiological signals) as well as more subjects. One common merit of BP4D-Spontaneous and BP4D+ is that they both provide expert FACS labels~\cite{facs_ekman} which are very useful in emotion analysis. There are also some low resolution databases captured using the Kinect sensor~\cite{cam3d_kinect,vt_kfer_kinect} designed for capturing 3D dynamic spontaneous behaviours.

Despite numerous 3D/4D facial expression databases are now publicly available, none of them contains samples collected in different sessions that allow us to investigate the use of dynamic behaviour for biometric applications\footnote{To the best of our knowledge, the databases collected for biometric applications contain only static 3D faces \cite{DMCSv1,frgc1,photoface,3DRMA}, hence their use for analysis of facial motion in biometric application is limited.}. As a consequence, research on dynamic 3D face recognition has fallen behind with static 3D face recognition. Only a few works were proposed in the past decade~\cite{hmm_fr4d_journal,alashkar_4dfr_journal,hayat_4dfr_bu4dfe,canavan_4dfr_bu4dfe}, most of them used BU-4DFE database~\cite{bu4dfe} which is limited for biometric applications. Arguably, the main reason is the lack of publicly available high quality 4D databases with many recording sessions that can be used for face recognition/verification. Furthermore, as the commonly used databases~\cite{alashkar_4d_data,bu4dfe} contain only one recording session per subject, the generalization ability of the tested method is doubtful.

As a matter of fact, all the aforementioned databases of 3D expressive samples (a) capture each subject only once (i.e. one recording session), which prohibits the use in a biometric scenario, (b) contain only posed or spontaneous expressions but not both and (c) generally include a small number of subjects. 

In this work, we take a very significant step forward and propose the 4D Facial Behaviour Analysis for Security (4DFAB) database which includes 180 participants on 4 different recording sessions spanning a period of 5 years. 4DFAB database contains over 1.8 million high resolution 3D facial scans and has been collected from 2012 to 2017. We believe that 4DFAB is an invaluable asset for many different tasks such as 3D/4D face and facial expression recognition using posed/spontaneous behaviours, building high quality expressive blendshapes, as well as synthesizing 3D faces for training deep learning systems. To better compare the proposed database with existing 4D face databases, we give an overview in Table~\ref{table:related_3d_databases}. 

To summarise, our contributions are:
\begin{itemize}
\item We present a database (we refer as \textbf{4DFAB}) of 180 subjects collected over a period of 5 years under four different sessions, with over 1,800,000 3D faces.
\item Our database contains both posed and spontaneous facial behaviours. The spontaneous behaviours were elicited by displaying stimuli that could elicit a variety of behaviours (\eg, from smile and laughter to cries and confusion). 
\item We investigate, for the first time, the use of spontaneous 4D behaviour for biometric applications~\footnote{The study in~\cite{benedikt2010assessing} only studied posed speech related and speech unrelated facial behaviour for biometric applications.}.
\item We demonstrate that expression blendshapes learned from our database are much more powerful than the off-the-shelf blendshapes provided by FaceWarehouse~\cite{facewarehouse}.
\end{itemize}
%
\begin{figure*}[tbh]
    \begin{subfigure}[t]{0.56\linewidth}
        \centering
        \includegraphics[width=3.7in,height=2.5in]{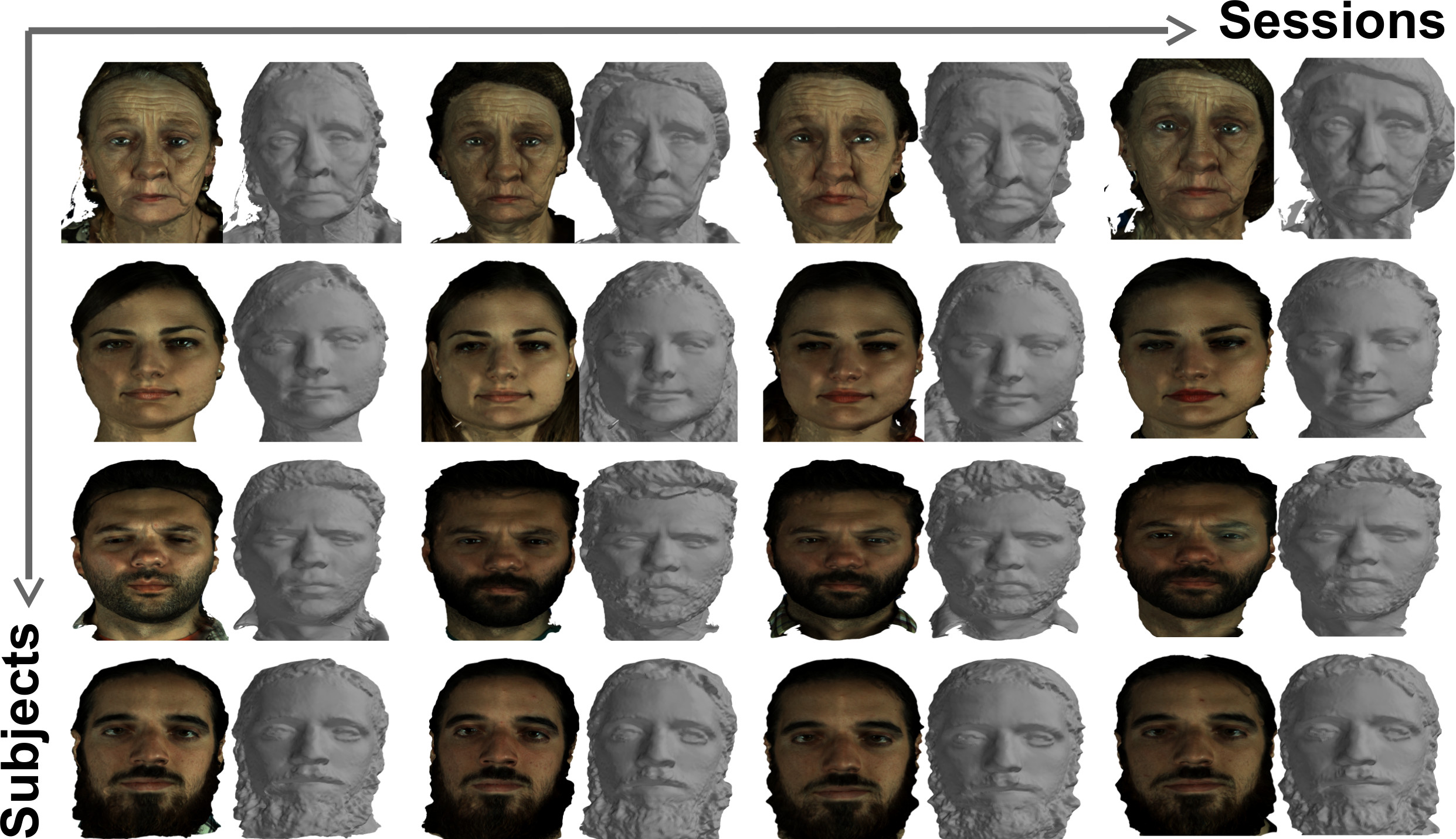}
        \caption{Examples of subjects in four sessions.}
    \end{subfigure}
    \hfill
    \begin{subfigure}[t]{0.43\linewidth}
        \centering
        \includegraphics[width=2.8in,height=2.4in]{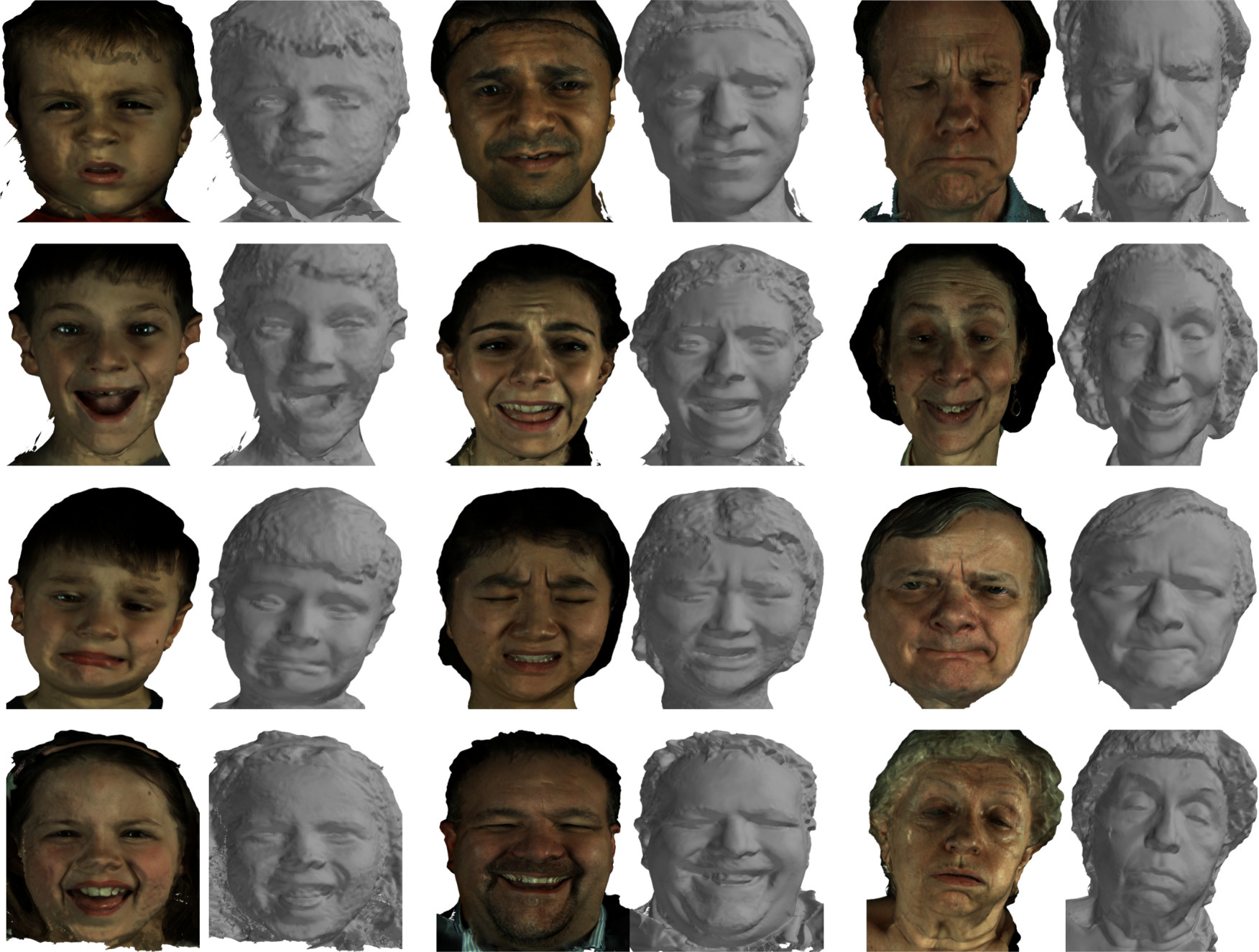}
        \caption{Examples of spontaneous expressions.}
    \end{subfigure}
\caption{Examples of 3D faces in the database.}
\label{fig:examples_orig}
\end{figure*}
\section{Data Acquisition}
For the past five years, we have collected a comprehensive dynamic facial expression database (4DFAB) that can be used for 3D face modeling, 3D face and expression recognition, etc. In this section, we will provide the details of this database.

\subsection{Capturing system setup}
We used the DI4D dynamic capturing system~\footnote{\url{http://www.di3d.com}} to capture and build 4D faces. This capturing system mainly consists of six cameras (two pairs of cameras and one pair of texture cameras, 60FPS, 1200x1600). The distance between the subject and camera plane is 140cm. Calibration was performed before every recording session, using a 10x10 20mm checkerboard. Two 4-lamp fluorescent lights were placed on each side to provide consistent and uniform lights. The complete set up is shown in Figure~\ref{fig:capturing_system}. Additionally, we added one microphone to record audio signal, one frontal grayscale camera (60FPS, 640x480) to record frontal face image, and a Kinect to record RGB-D data (30FPS, 640x480). They were synchronized with the 4D recording using trigger from the DI4D capturing system, and will be publicly available after the initial release of high resolution 4D faces. Nevertheless, we mainly focus on the 4D data in this paper.
\begin{table*}
\centering
\begin{tabular}{@{}lll@{}}
\toprule
\textbf{Task} & \textbf{Activity} & \textbf{Target Emotion} \\ \midrule
T1.1 & \makecell[lt]{Task: perform 6 expressions (anger, disgust, fear, happiness, sadness, surprise).} & N/A \\
T1.2 & \makecell[lt]{Task: word utterances (puppy, baby, mushroom, password, ice cream, bubble,\\Cardiff, bob, rope), three times each.} & N/A \\
T1.3 & Video clip: jump scare immediately after words reading, finish with a joke. & Surprise, fear, happiness. \\
T1.4 & Task: count backwards from 1000 by 7s. & Embarrassment, nervousness.\\
T1.5 & Video clip: two disgusting and fearful clips on human eyes. & Disgust, fear, surprise. \\
T1.6 & Video clip: comedy \textit{Women: Know Your Limits!} & Happiness, anger. \\
T1.7 & Video clip: funny moments of cat. & Happiness. \\
T1.8 & Video clip: funny dance by a parrot. & Happiness. \\ \bottomrule
\end{tabular}
\caption{Tasks of Session 1 recording.}
\label{tab:session1}
\end{table*}
\begin{table*}
\centering
\begin{tabular}{@{}lll@{}}
\toprule
\textbf{Task} & \textbf{Activity} & \textbf{Target Emotion} \\ \midrule
T2.1 & Task: perform 6 basic expressions (same as T1.1). & N/A \\
T2.2 & Task: word utterances (same as T1.2). & N/A \\
T2.3 & Task: add 1 to each digit in the given numbers. & Embarrassment, nervousness. \\
T2.4 & Task: add 3 to each digit in the given numbers. & Embarrassment, nervousness. \\
T2.5 & Video clip: jump scare concealed in a relaxing video. & Surprise, fear. \\
T2.6 & Video clip: a collection of disgusting but funny movie scenes. & Disgust, happiness. \\
T2.7 & Video clip: emotional Thai story. & Sadness. \\
T2.8 & Video clip: clip from the movie \textit{Up}. & Sadness. \\
T2.9 & Video clip: Jimmy Kimmel's \textit{Halloween Candy Prank}. & Happiness. \\ \bottomrule
\end{tabular}
\caption{Tasks of Session 2 recording.}
\label{tab:session2}
\end{table*}
\begin{table*}
\centering
\begin{tabular}{@{}lll@{}}
\toprule
\textbf{Task} & \textbf{Activity} & \textbf{Target Emotion} \\ \midrule
T3.1 & Task: perform 6 basic expressions (same as T1.1). & N/A \\
T3.2 & Task: word utterances (same as T1.2). & N/A \\
T3.3 & Video clip: jump scare concealed in a color blindness test video. & Surprise, fear. \\
T3.4 & Task: subtract 1 to each digit in the given numbers. & Embarrassment, nervousness. \\
T3.5 & Task: subtract 3 to each digit in the given numbers. & Embarrassment, nervousness. \\
T3.6 & Video clip: man eating giant larva from \textit{Man vs. Wild}. & Disgust. \\
T3.7 & Video clip: Derek Redmond's emotional Olympic story in 1992. & Sadness. \\
T3.8 & Video clip: funny fails compilation. & Happiness, surprise. \\
T3.9 & Video clip: emotional shadow dance. & Sadness. \\ \bottomrule
\end{tabular}
\caption{Tasks of Session 3 recording.}
\label{tab:session3}
\end{table*}
\begin{table*}
\centering
\begin{tabular}{@{}lll@{}}
\toprule
\textbf{Task} & \textbf{Activity} & \textbf{Target Emotion} \\ \midrule
T4.1 & Task: perform 6 basic expressions (same as T1.1). & N/A \\
T4.2 & Task: word utterances (same as T1.2). & N/A \\
T4.3 & Video clip: jump scare immediately after words reading. & Surprise, fear. \\
T4.4 & Task: replace the character with corresponding number in the given strings. & Embarrassment, nervousness. \\
T4.5 & Video clip: disgusting but funny prank. & Disgust, happiness. \\
T4.6 & Video clip: funny parody on iPhone. & Happiness. \\
T4.7 & Video clip: edited commercial \textit{Mind the Gap}. & Sadness. \\ \bottomrule
\end{tabular}
\caption{Tasks of Session 4 recording.}
\label{tab:session4}
\end{table*}
\subsection{Experiment design and emotion elicitation}
Our experiments aim at capturing posed expressions, spontaneous expressions and any other evident facial movements. We define posed expression as the participant deliberately making the expression that has the same semantic meaning as the target expression. Spontaneous expression, as its name suggests, is the natural and spontaneous emotion shown by the participant during the experiment. Video clip viewing was our main way to elicit such expressions. Except for expressions, we collected some facial movements that might not correspond to any emotions, examples of which included flaring nostrils, biting lip, raising eyebrow, etc.

Each participant was asked to read and sign a consent form, which allowed the use of data for research purposes. Before the experiment, participant was asked to take off any glasses, hat or scarf, and wear hairnet/hairband to prevent occlusion if necessary. After that, we calibrated the cameras, adjusted the seat, and made a preview capture. In order to acquire high-fidelity captures, participant was asked to avoid large body and head movements during the recording.

Within each recording session, we first asked the participant to perform 6 basic expressions (i.e. anger, disgust, fear, happiness, sadness and surprise) and pronounce the nine words (i.e. puppy, baby, mushroom, password, ice cream, bubble, Cardiff, bob, rope) three times in order. These two tasks were repeated for every session. Then, a few tasks involving words and numbers were undertaken. After this, we showed the participant several videos to elicit spontaneous expressions. Between two successive tasks, we gave the participant a short break (a minute or two) to reset his/her emotion state. 

To provide 4D data for face recognition/verification purposes, we created 4 recording sessions with different video stimulus, and invited the same participant to attend 4 times. We list the tasks of each recording session of the proposed database in Table~\ref{tab:session1}~\ref{tab:session2}~\ref{tab:session3}~\ref{tab:session4}, which describe Session 1, 2, 3 and 4 respectively. Note that there might be multiple target emotions in one task, as we are anticipating different reactions from different subject. Examples of the same subject in different sessions are shown in Figure~\ref{fig:examples_orig}. We recorded over 40 hours of raw data, however, it was neither necessary nor feasible to reconstruct all of them. Therefore, we browsed every sequence and manually divided it into different segments of facial expression or movements. The final database included 1.8 million 3D meshes (equivalent of 8.4 hours recording), and took more than 20TB of storage space.
\begin{figure*}[tbh]
    \begin{subfigure}[t]{0.68\linewidth}
        \centering
        \includegraphics[height=1.75in]{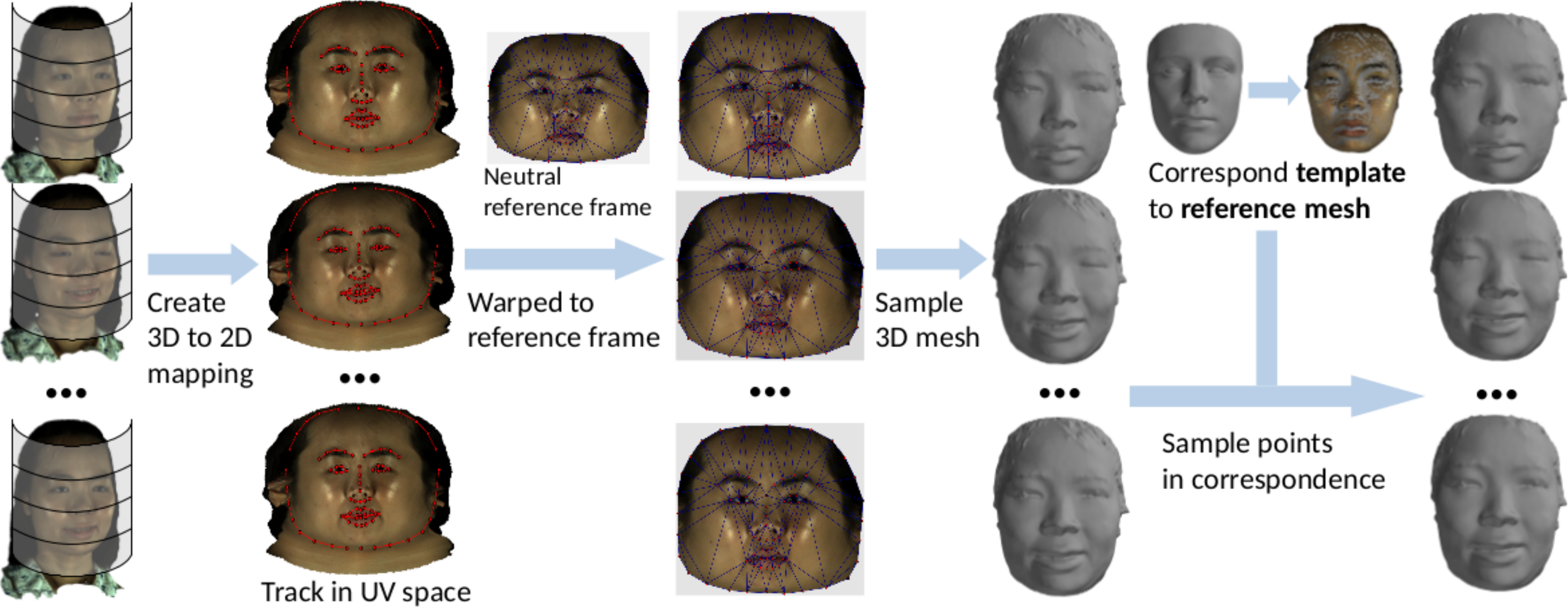}
        \caption{4D alignment pipeline.}
    \end{subfigure}
    \hfill
    \begin{subfigure}[t]{0.31\linewidth}
        \centering
        \includegraphics[height=1.75in]{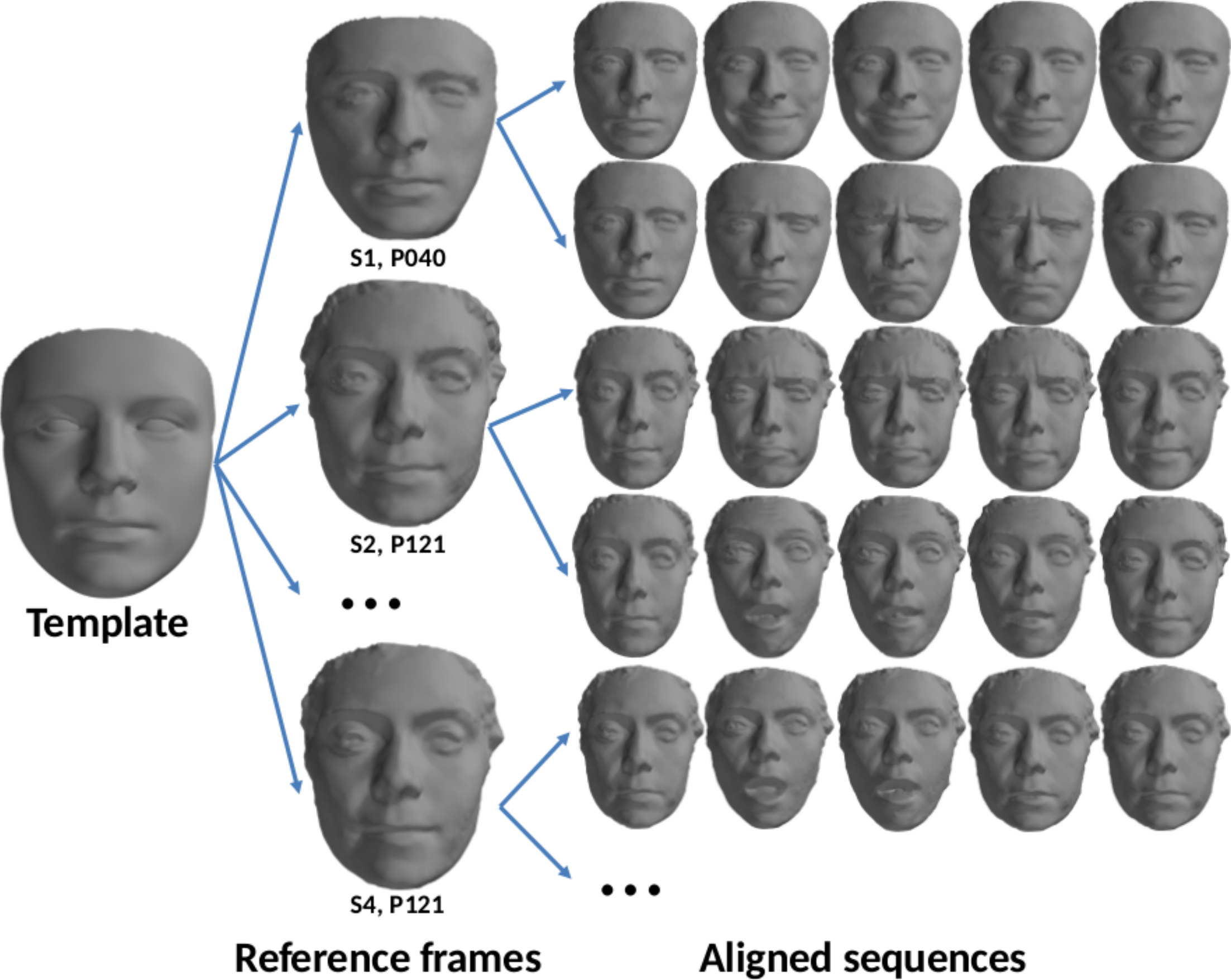}
        \caption{Overview of correspondences.}
    \end{subfigure}
\caption{Our framework for establishing 4D dense correspondences.}
\label{fig:framework_4d}
\end{figure*}
\subsection{Participants}
180 participants (60 females, 120 males) aged from 5 to 75 took part in the experiment. The majority of them were recruited from our institute's administrative section and departments (Engineering, Business, Medicine, etc.), the other subjects (over 40) were volunteers from outside the college. They are from over 30 different culture backgrounds including Chinese, Greek, British, Spanish, etc. Ethnicity includes Caucasian (Europeans and Arabs), Asian (East-Asian and South-Asian) and Hispanic/Latino. The distribution of ages (based on the first attendance) and ethnic groups are summarised in Table~\ref{tab:age_ethnic_dist}. 

Among all the participants, 179 subjects attended the first session, 100 subjects participated the second session, while 81 and 75 participants have come for the third and fourth time respectively. The average time interval between two consecutive attendances is 219 days (shortest: 1 day, longest: 1,654 days). Among them, 56\% are recorded within 3 months, 23\% are between 3 to 12 months and 21\% for over a year.

\begin{table}[tbh]
\begin{subtable}[b]{0.48\linewidth}
    \small
    \centering
    \begin{tabular}{ccc}
    \hline
    Age & Num. & Prop.\\ 
    \hline
     5-18 & 5   & 2.8\%\\
    19-29 & 115 & 63.9\%\\
    30-39 & 41  & 22.8\%\\
    40-49 & 9   & 5.0\%\\
    over 50 & 10& 5.5\%\\ \hline
    \end{tabular}
    \caption{Age distribution}
\end{subtable}
\hfill
\begin{subtable}[b]{0.5\linewidth}
    \small
    \centering
    \begin{tabular}{ccc}
    \hline
    Ethnic & Num. & Prop.\\ 
    \hline
    Caucasian & 101 & 56.1\%\\
    Asian & 63 & 35.0\%\\
    \makecell[lt]{Hispanic/\\Latino} & 16 & 8.9\%\\ \hline
    \end{tabular}
    \caption{Ethnicity distribution}
\end{subtable}
\caption{Distribution of age and ethnicity in our database.}
\label{tab:age_ethnic_dist}
\end{table}
\subsection{Data processing and organization}
Six synchronised 2D video sequences were recorded during experiment. For every pair of stereo images within the sequence, a passive stereo photogrammetry method was employed to produce a range map which was subsequently used for reconstructing 3D face. Ten machines were actively running for 1.5 years to reconstruct nearly two million selected frames. A summary of reconstructed data is given in Table~\ref{tab:summary}. The vertex number of reconstructed 3D meshes ranges from 60k to 75k, with the maximum edge length allowed in mesh being 2mm.
\begin{table}
\centering
\begin{tabular}{ccccc}
\hline
Session & Participants & Frames & Avg. frames\\ 
\hline
S1 & 179 & 768,290  & 4,292\\
S2 & 100 & 409,272  & 4,093\\
S3 & 81  & 345,921  & 4,271\\
S4 & 75  & 312,030  & 4,160\\ 
\hline
\end{tabular}
\\
Total number of frames = 1,835,513
\caption{Summary of reconstructed 4D data.}
\label{tab:summary}
\end{table}
%
%
\begin{figure*}[tbh]
\includegraphics[width=0.995\linewidth,height=0.4\linewidth]{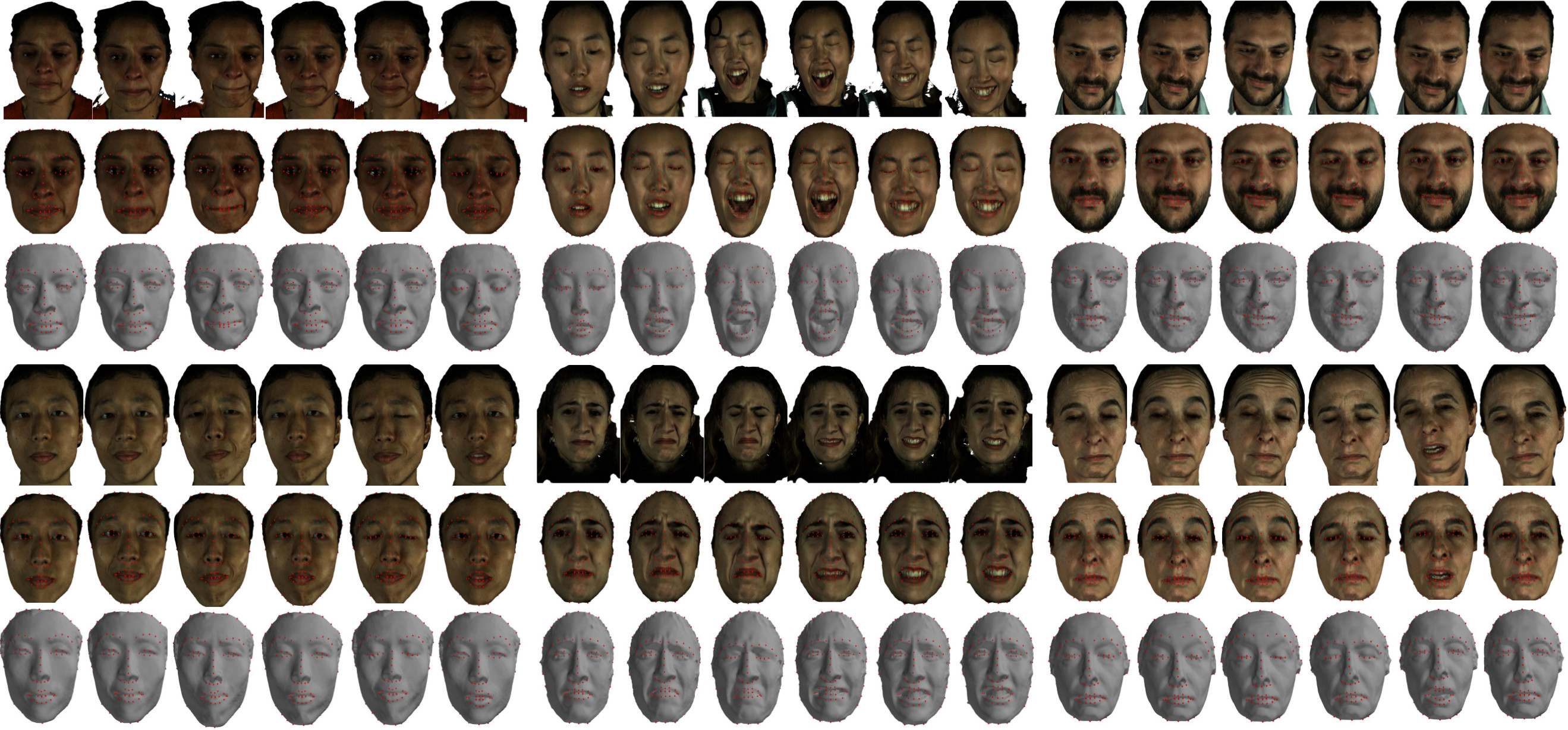}
\caption{4D sequences that are in full correspondence with template, displayed with 79 landmarks. For each sequence, top row shows original scans, middle and bottom rows are the registered 3D meshes with and without texture respectively.}
\label{fig:visualise_4d}
\end{figure*}
\section{Establishing 4D Dense Correspondences}
It is very important to establish dense correspondences between every mesh and an universal template. There are two popular approaches for this, one is through non-rigid image registration in UV-space~\cite{3dmm_revisited_PatelS09, Cosker_iccv11}, another is directly aligning the template to the target mesh (\eg using NICP~\cite{nonrigid_icp,Booth2017,anicp}). They both provide accurate correspondences, whereas UV-based approaches are more powerful and computationally efficient (refer to~\cite{Booth2017} for an in-depth comparison). Although some intricate face parts (\eg interiors of nostrils, ears) are precluded from the UV map, it should not affect our data which do not exhibit those details. In this section, we explain our UV-space-based alignment framework (also demonstrated in Figure~\ref{fig:framework_4d}).

\subsection{2D to 3D mapping in UV space}
Firstly, we create a 2D to 3D mapping by a bijective mapping from 2D positions in UV space to the corresponding 3D point in the mesh (see Figure~\ref{fig:framework_4d}(a)). Assume that such mapping could accurately represent a 3D face, establishing dense correspondence between any two UV images will automatically return us a dense 3D-to-3D correspondence for their corresponding 3D meshes. This is beneficial because it transfers the challenging 3D registration problem to the well-solved 2D non-rigid image alignment problem. Furthermore, in our specific case where 1.8 million meshes need to be aligned, this is obviously more reliable and computationally efficient. We employ an optimal cylindrical projection method~\cite{james_uv} to synthetically create a UV space for each mesh, and produce a UV map $I$ with each pixel encoding both spatial information (X, Y, Z) and texture information (R, G, B), on which we perform non-rigid alignment.
\begin{figure}
\centering
\includegraphics[width=0.4\linewidth]{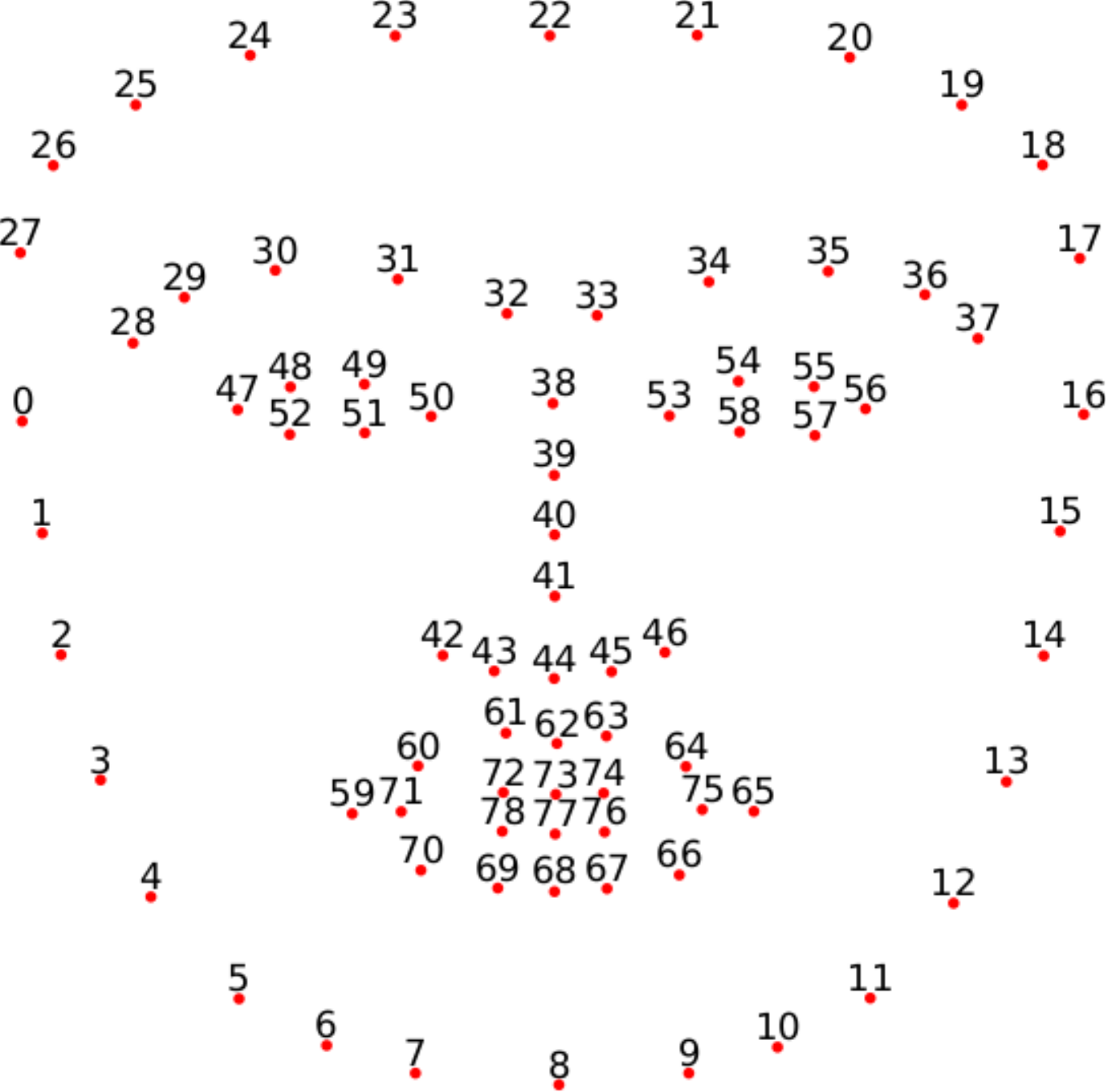}
\hspace{25pt}
\includegraphics[width=0.35\linewidth, height=0.40\linewidth]{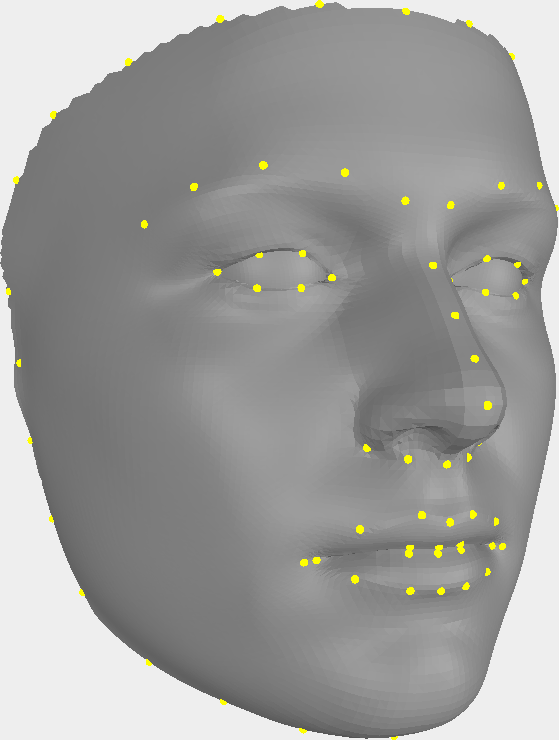}
\caption{Definition of 79 facial landmarks (left) and exemplar of annotated template (right).}
\label{fig:definition_lm79}
\end{figure}
\subsection{Non-rigid UV image alignment and sampling}
Several non-rigid alignment methods in UV space have been proposed. One way (denoted as UV-OF) is applying Optical Flow on the UV texture and the 3D cylindrical coordinates to align two UV maps~\cite{3dmm_vetter, Bradley_tog_2011}. Another approach utilises key landmarks fitting and Thin Plate Spline (TPS) warping~\cite{3dmm_revisited_PatelS09, Cosker_iccv11} (referred as UV-TPS). 
We follow the UV-TPS approach because UV-OF might produce drift artifacts as the optical flow tracking continues, while 2D landmarks detection usually provides stable and consistent facial landmarks to avoid drifting. 

Compared with~\cite{Cosker_iccv11}, we made several changes to suit our data. Firstly, we built session-and-person-specific Active Appearance Models (AAMs)~\cite{alabortIJCV2016,akshay_pami} to automatically track feature points in the UV sequences. This means that 4 different AAMs would be built and used separately for one subject. Main reasons behind this are (1) textures of different sessions differ due to several facts (i.e. aging, beard, make-ups, experiment lighting condition, etc.); (2) person-specific model is proven more accurate and robust in specific domain~\cite{psm_2012}. Therefore, we manually selected 1 neutral and 2-3 expressive meshes per person and session, and annotated 79 3D landmarks (Figure~\ref{fig:definition_lm79}) using a 3D landmarking tool~\footnote{\url{https://www.landmarker.io}}~\cite{menpo14}. Overall, 435 neutral meshes and 1047 expression meshes were labelled. They were unwrapped and rasterised to UV space, and then grouped for building the corresponding AAMs. Note that we flipped each UV map to increase fitting robustness.

Once all the UV sequences were tracked with 79 landmarks, they were then warped to the corresponding reference frame using TPS, and thus achieving the 3D dense correspondence. For each subject and session, we built one specific reference coordinate frame from his/her neutral UV map. From each warped frame, we could uniformly sample the texture and 3D coordinates. Eventually, a set of non-rigidly corresponded 3D meshes under the same topology and density were obtained. Here, an extra rigid alignment step might be performed to further remove similarity differences.

\subsection{Establishing correspondence to face template}
Given that meshes have been aligned to their designated reference frame, the last step is to establish dense 3D-to-3D correspondences between those reference frames and a 3D template face. This is a 3D mesh registration problem, and can be solved by Non-rigid ICP~\cite{nonrigid_icp}. NICP extends the ICP algorithm by assuming local affine transformation for each vertex, and iteratively minimises the distance between source and target meshes with adjustable stiffness constraint. We employed it to register the neutral reference meshes to a common template - Basel mean face~\cite{bfm09}. We did not use the full Basel face, because our meshes might not fully describe ears, neck and nostrils. Thus we crop the original face and flatten the nostrils to get a new template (see Figure~\ref{fig:definition_lm79} for the modified template). Upon completion of this step, we corresponded every 3D mesh to one single template, an overview of correspondences is depicted in Figure~\ref{fig:framework_4d}(b). We also plot many example sequences in Figure~\ref{fig:visualise_4d}.

\section{Building Expression Blendshape Model}
\label{sec:blendshape}
In order to build the blendshapes we used the methodology proposed in~\cite{local_blendshape}. In particular, we annotated the apex frames (frames with maximum facial change) of all the pose expression sequences (anger, disgust, happiness, fear, sadness and surprise). For each of the sequences we subtracted the neutral mesh of the sequence from the corresponding apex frame. In this manner, we created a set of $m$ difference vectors $\mathbf{d}_i \in \mathbb{R}^{3n}$ which were then stacked into a matrix $\mathbf{D}=[\mathbf{d}_1, ..., \mathbf{d}_m] \in \mathbb{R}^{3n \times m}$, where $n$ is number of vertices in our mesh. Afterwards a variant of sparse Principal Component Analysis (PCA) was applied to our data matrix $\mathbf{M}$ to identify sparse deformation components (the interested reader may refer to~\cite{local_blendshape} for more details). Note that for visualization purposes, we blended all our meshes to the template provided by~\cite{Booth2017} to recover a face with ears and neck. We will provide both quantitative and qualitative measures to evaluate our expression blendshape in the next section.
\begin{figure*}[t!]
\includegraphics[width=\linewidth]{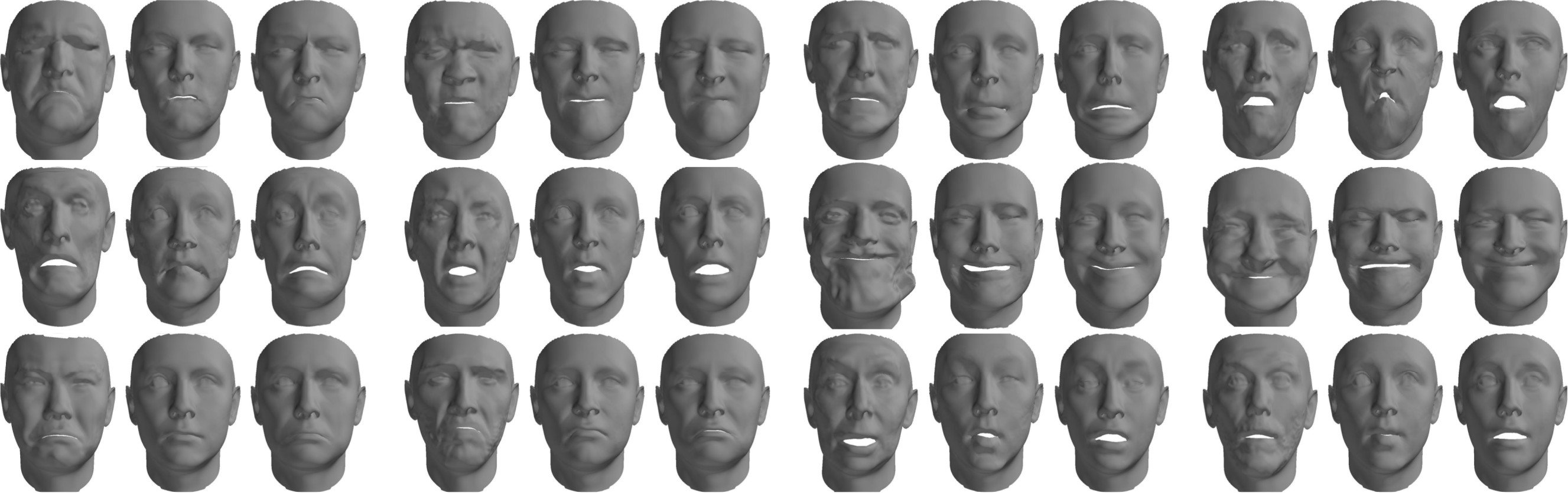}
\centering
Scan \hspace{0.035\linewidth} FW \hspace{0.04\linewidth} \textbf{Ours}
\hspace{0.045\linewidth} Scan \hspace{0.04\linewidth} FW \hspace{0.035\linewidth} \textbf{Ours} \hspace{0.045\linewidth} 
Scan \hspace{0.035\linewidth} FW \hspace{0.04\linewidth} \textbf{Ours}
\hspace{0.045\linewidth} Scan \hspace{0.04\linewidth} FW \hspace{0.035\linewidth} \textbf{Ours}
\caption{Comparison of FaceWarehouse blendshape model~\cite{facewarehouse} and our expression blendshape model. Note that we only transfer the expression to the mean face of~\cite{Booth2017}, not involving any modeling of face identity.}
\label{fig:blendshapes}
\end{figure*}
%
\section{Experiments}
We conducted several experiments using the densely corresponded sequences, and report the performances as the baselines of our 4DFAB database for the research community. 

\subsection{Facial expression recognition}
We did two standard FER experiments on 6 posed expressions, one was static and another was dynamic. Considering that not all the sessions had full attendance, separate subject-independent recognition experiments were set up. Within each session, we created a 10-fold partition, every time one fold was used for testing, the others were used for training. Both static and dynamic experiments used the same 10-fold partition.

\subsubsection{3D Static expression recognition}
We manually labelled the apex frames of each expression sequence. Because the apex interval varied from sequences, we trimmed the apex period and generated a balanced set that had 5 meshes per subject per session. We rasterised every 3D mesh into Z-buffer, and extracted the main face regions (covering eyes, mouth, cheeks and nose) based on 79 facial landmarks. The region was further divided into non-overlapping blocks, for which Histogram of Oriented Normal Vectors (HONV) \cite{honv} were computed. After this, PCA and LDA were used for dimensionality reduction, a multi-class SVM~\cite{svm} was employed to classify expressions. Radial Basis Function (RBF) kernel was selected, whose parameters were chosen by an empirical grid search. 
We achieved a recognition rate of 70.27\% Session 1 experiment, 69.02\%, 66.91\% and 68.89\% in Session 2, 3 and 4 experiments respectively.
\begin{table}
\centering
\begin{tabular}{cccccc}
    \hline
    Method & S1 & S2 & S3 & S4\\
    \hline
    Ours   & 72.69 & 68.88 & 71.1 & 70.09\\
    FW~\cite{facewarehouse} & 71.02 & 68.03 & 68.35 & 66.81\\
    \hline
\end{tabular}
\caption{Recognition Rates (RR) [\%] obtained from 3D Dynamic facial expression recognition experiments.}
\label{tab:3d_dynamic_fer}
\end{table}
\subsubsection{3D Dynamic expression recognition}
\label{sec:3d_dynamic_fer}
We used Long Short-term Memory (LSTM)~\cite{lstm} to recognise dynamic expressions. For every expressive 3D face, we computed its facial deformation with regard to the corresponding neutral face. We then projected it to our blendshape model as well as the FaceWarehouse (FW) model~\cite{facewarehouse} to obtain the sparse representations of expression, which would be used as the feature. In order to reduce noises, Kalman filter~\cite{kalman_filter} was further applied to each dimension of features within the segments. For each experiment, only one standard LSTM layer was utilised, whose capacity was empirically decided according to the number of available training data. The Adam algorithm~\cite{adam} was selected, with a learning rate of 0.001, batch size 12, and 15 epochs at max. Results in Table~\ref{tab:3d_dynamic_fer}(b) showed that even with the simplest feature (i.e. blendshape parameters) and a basic LSTM network, we could achieve around 70\% in recognition rate, which suggested that our 4D alignment was quite accurate and reliable. Moreover, recognition performances of our blendshape were better than FW in every session, which showed that our blendshape could model expression more accurately.

\subsection{3D Dynamic face recognition}
We report the results for 3D dynamic face recognition using 6 basic expressions respectively. We selected 74 subjects who have attended four sessions and performed all expressions. For each experiment (namely each expression), we performed a leave-one-out cross-validation - each time one session was left out for testing. The same feature, temporal filtering and LSTM network from Section~\ref{sec:3d_dynamic_fer} were employed for these tasks, except that LSTM capacity was decided experimentally. Both our expression blendshape model and FW were tested. Although our blendshape (named as \textbf{3DMM-exp}) models only the facial deformation and does not involve shape information of identity, it is essentially a variant of 3DMM. Therefore, it is interesting to compare it the standard 3DMM that models both expression and identity information. We built a 3DMM using 1482 aligned meshes with ground-truth 79 landmarks, and projected all the meshes to this model to obtain the shape parameters. We empirically selected the first 68 components and used it as feature descriptor (denoted as \textbf{3DMM} in Table~\ref{tab:3d_dynamic_fr}). 

From Table~\ref{tab:3d_dynamic_fr}, not surprisingly, using 3DMM we could recognise 96\% of the test instances on average. Since 3DMM jointly models the identity and expression, its shape parameters embed information from both sides, thus lowers the difficulty for LSTM to recognise test subject. Nevertheless, with our expression blendshape model in which identity is not present, we could achieve meaningful performances using anger, disgust and happiness (66.22\%, 66.55\% and 67.23\% respectively). To the best of our knowledge, we are the first one to exploit dense shape deformation (without explicit identity information) in 3D dynamic face recognition. Our results also indicate that the use of dynamic expression sequences in a biometric scenario is worth investigating.
\begin{table}
\centering
\small
\begin{tabular}{lcccccc}
    \hline
    Method & AN & DI & FE & HA & SA & SU\\ 
    \hline
    FW~\cite{facewarehouse} & 45.61 & 51.69 & 41.89 & 54.73 & 45.95 & 49.66\\
    3DMM-exp & 66.22 & 66.55 & 62.84 & 67.23 & 59.46 & 61.15\\ 
    3DMM & 96.62 & 95.95 & 96.28 & 97.3 & 96.62 & 95.61\\ 
    \hline
\end{tabular}
\caption{Recognition Rates (RR) [\%] obtained from 3D Dynamic face recognition experiments using 6 expressions (AN-Anger, DI-Disgust, FE-Fear, HA-Happiness, SA-Sadness, SU-Surprise).}
\label{tab:3d_dynamic_fr}
\end{table}
\subsection{3D Face verification}
Two verification experiments with posed expressions and spontaneous smile respectively were undertaken. We borrowed the verification methods from~\cite{verification_stefanos}, in which the facial deformation was calculated between the neutral and expression apex frame, and used for verification. Supervised and unsupervised dimensionality reduction techniques were applied to extract sparse feature. In our cases, deformation was computed as the difference between aligned expressive mesh and its neutral mesh.
\begin{figure}[tbh]
\begin{subfigure}[t]{0.51\linewidth}
    \centering
    \includegraphics[width=\linewidth]{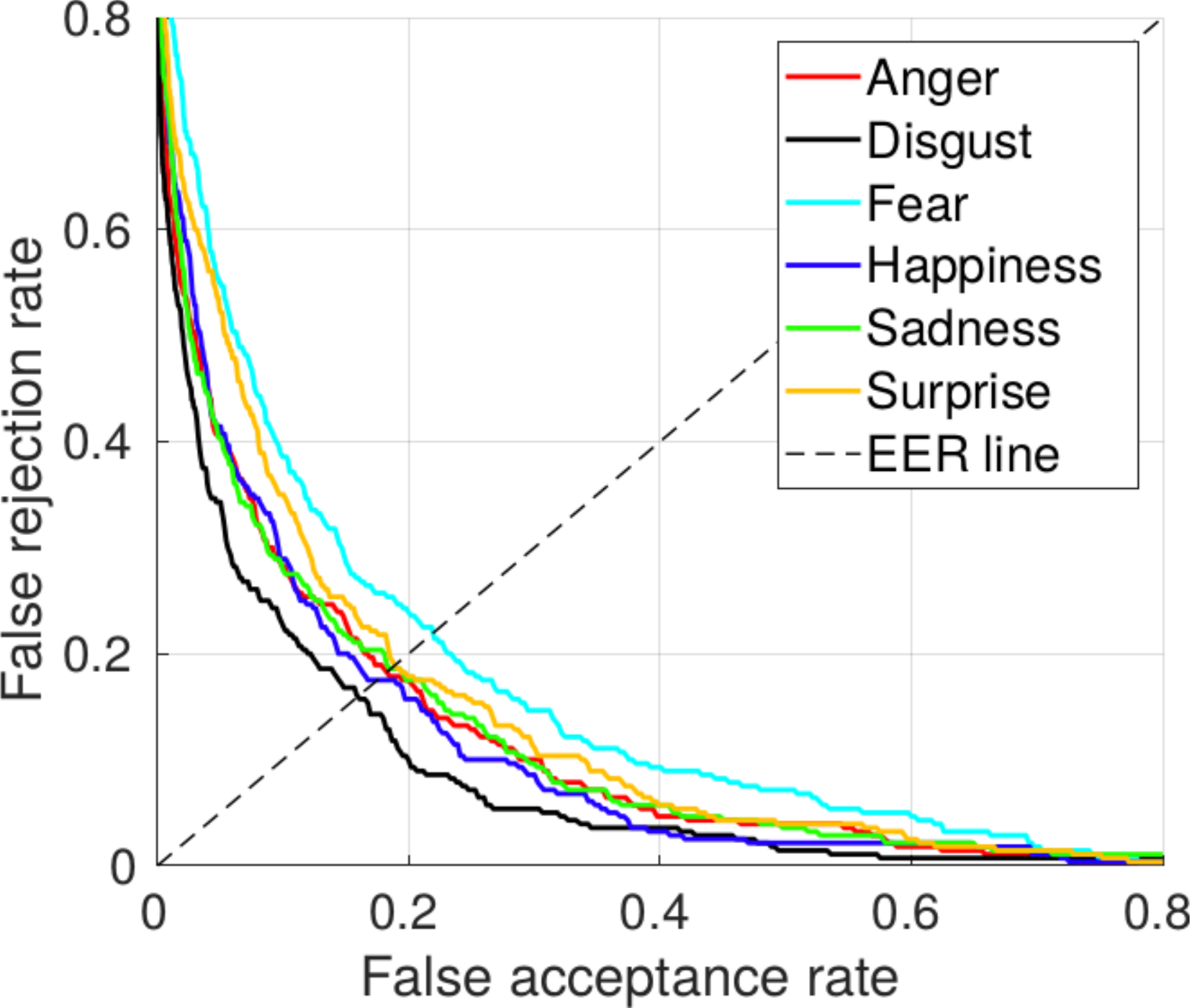}
    \caption{\footnotesize{Posed expressions}}
\end{subfigure}
\hfill
\begin{subfigure}[t]{0.46\linewidth}
    \centering
    \includegraphics[width=\linewidth]{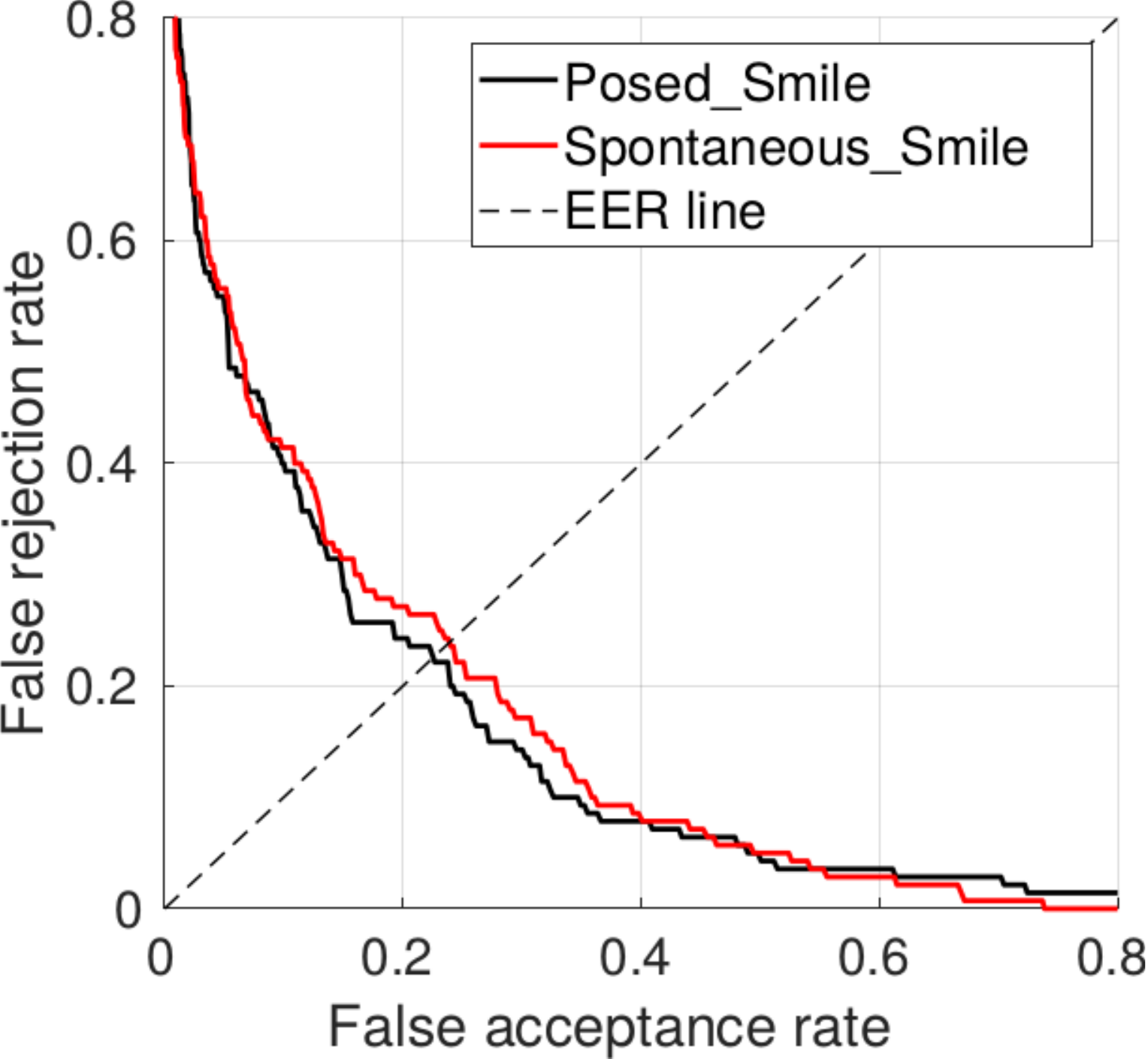}
    \caption{\footnotesize{Spontaneous/posed smile}}
\end{subfigure}
\caption{The ROC curves for (a) Posed expressions and (b) Spontaneous and posed smile.}
\label{fig:3d_fveri}
\end{figure}
\subsubsection{Verification using posed expressions}
In this experiment, each posed expression was tested separately. Based on 4 recording sessions, 4 experimental sessions were implemented by employing the leave-one-out scheme and rotation estimates. For all the experiments, we excluded 131 subjects who did not attend all sessions and used them as test impostor claims. For each experiment, the leave-out session would be used as the test set (genuine claims), thus the other three sessions were used for training. The number of genuine claims was 70. We used only one apex frame per client. The rest training and testing procedures were identical as those described in~\cite{verification_stefanos}. We showed the verification performance in the Receiver Operating Characteristic (ROC) curve with False Rejection Rate (FRR) and False Acceptance Rate (FAR). The performance of a verification system is often quoted by a particular operating point of the ROC curve where FAR = FRR, which is called Equal Error Rate (EER)~\cite{ZAFEIRIOU20072798}.

The curves are plotted in Figure~\ref{fig:3d_fveri}(a), EER for anger, disgust, fear, happiness, sadness, surprise are 18.2\%, 15.9\%, 21.9\%, 17.5\%, 18.5\% and 18.8\% respectively. It suggests that, for our posed expression data, disgust (15.9\%) and happiness (17.5\%) are more informative than the others in verification.

\subsubsection{Verification using spontaneous smile}
To demonstrate the usefulness of our spontaneous data, we used the apex frame of spontaneous smile/laughter per subject and session for verification. The protocol was similar to previous verification experiment. Four sessions were implemented with the leave-one-out scheme. For the impostors, we reserved 124 subjects who did not have a full set of smiles from all sessions. The number of genuine clients across all sessions was 39. We also applied the same protocol for the posed Happiness to compare. The ROC curves are plotted in Figure~\ref{fig:3d_fveri}(b). The EER achieved by posed smile was 22.6\%, while spontaneous smile was 23.9\%. The attained results indicat that spontaneous smile is as useful as its posed counterpart for automatic person verification. As far as we know, this is the first investigation on the use of 4D spontaneous behaviours in biometric application.

\subsection{3D Dynamic speech recognition}
We conducted a dynamic speech recognition experiment on nine words: puppy, baby, mushroom, password, ice cream, bubble, Cardiff, bob, rope. This experiment was performed session-wise due to the same reason as expression recognition. For each session, we created a 10-fold partition, with one fold for testing, and the rest nine folds for training. The same LSTM~\cite{lstm} network was utilised. Since we were only interested in the mouth part, we defined a mouth region in the face template and extract it from every mesh in the sequence. Similarly, we built a 3D mouth model using all the meshes with landmark annotation, and kept 98\% of the variations (46 components). All the meshes were then projected to this 3D mouth model to retain the shape parameters, which would be used as our feature descriptors. Kalman filter~\cite{kalman_filter} was applied to smooth the feature sequence. As shown in Table~\ref{tab:3d_dynamic_sr}, without any usage of texture or elaborated features, even our worse recognition performance (S4) is nearly 69\%, while the best performance (S1) is 77.89\%. This is quite likely contributing to an accurate dense alignment in mouth part.
\begin{table}
\centering
\begin{tabular}{cccccc}
    \hline
    Session & S1 & S2 & S3 & S4\\
    \hline
    LSTM & 400 & 350 & 300 & 300\\ 
    RR[\%] & 77.89 & 75.17 & 70.28 & 68.47\\
    \hline
\end{tabular}
\caption{Recognition Rates (RR) obtained from 3D Dynamic speech recognition experiments on 9 words utterances. LSTM capacity for each session was also reported.}
\label{tab:3d_dynamic_sr}
\end{table}

\subsection{Evaluation of expression blendshape model}
\begin{figure}
\centering
\includegraphics[width=0.8\linewidth]{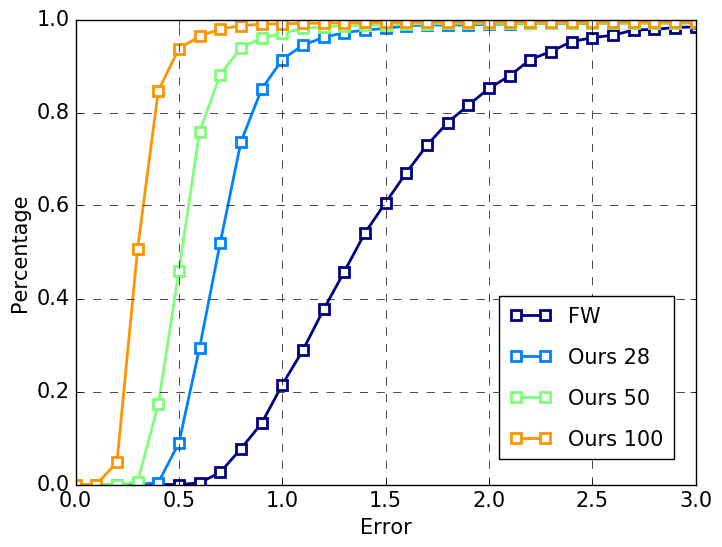}
\caption{Cumulative reconstruction errors achieved with different blendshapes over randomly selected spontaneous expressions from our database.}
\label{fig:reconstruction_figure}
\end{figure}
\begin{figure*}[tbh]
    \begin{subfigure}[t]{0.48\linewidth}
        \centering
        \includegraphics[width=0.95\linewidth]{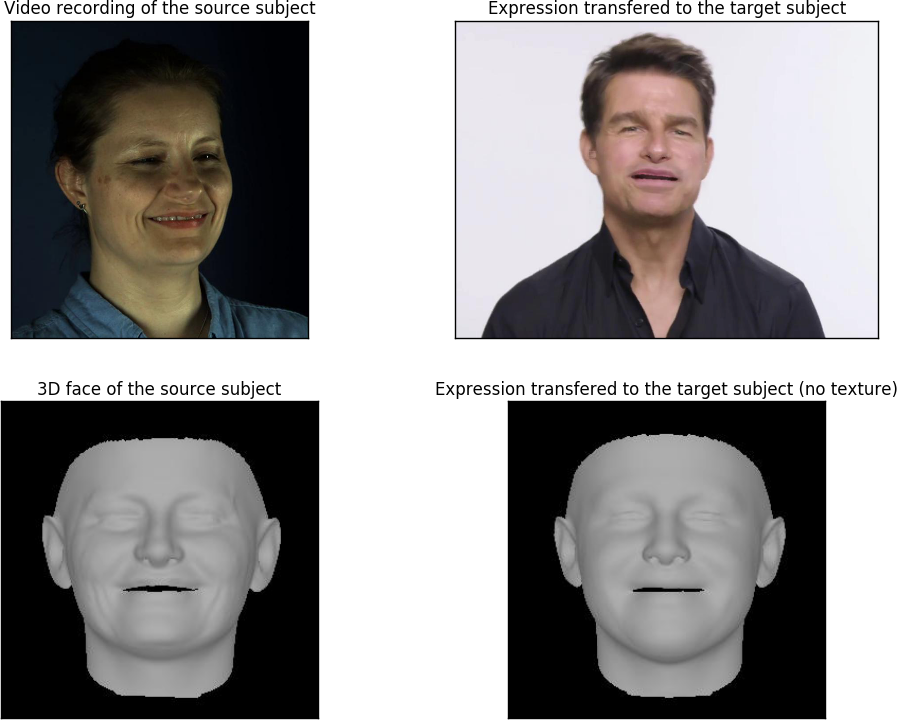}
        \caption{Expression transfer given a spontaneous expression sequence from subject P085 of Session 4.}
    \end{subfigure}
    \hfill
    \begin{subfigure}[t]{0.48\linewidth}
        \centering
        \includegraphics[width=0.95\linewidth]{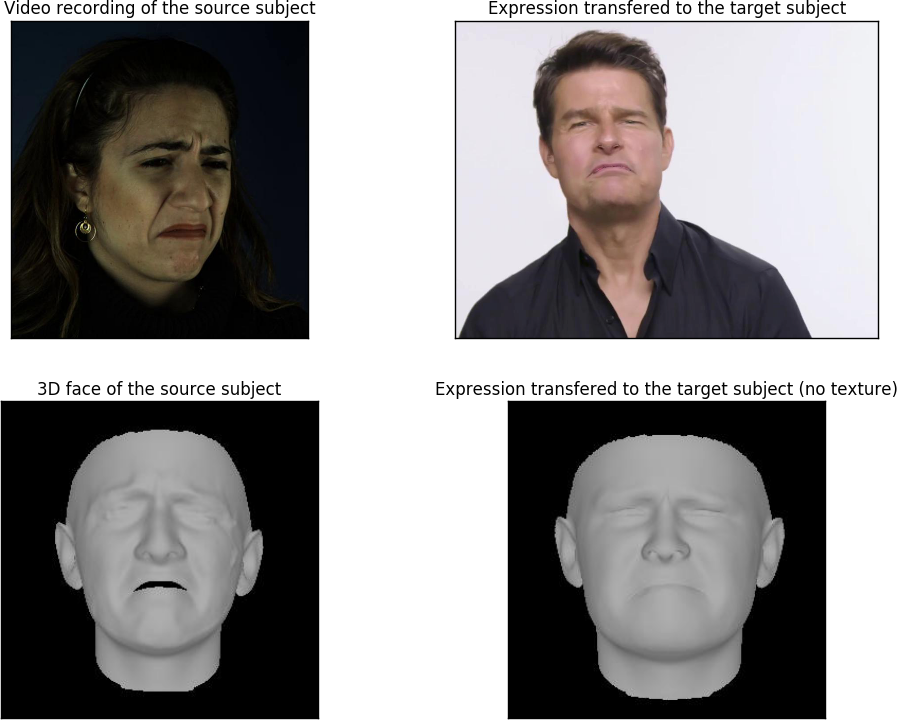}
        \caption{Expression transfer given a spontaneous expression sequence from subject P163 of Session 4.}
    \end{subfigure}
    \caption{Screenshots of expression transfer demo video.}
    \label{fig:expression_transfer}
\end{figure*}
We compare our blendshape model with FaceWarehouse (FW) model~\cite{facewarehouse} in spontaneous expressions reconstruction. We randomly selected 1453 frames that display spontaneous expressions, from which, we computed the facial deformation in the same way as described in~\ref{sec:blendshape} and reconstructed it using both blendshape models. We calculate the reconstruction error and plot the cumulative error curves for models with different number of expression components in Figure~\ref{fig:reconstruction_figure}. To provide a fair comparison with FW, we report the performance of our model (\textbf{Ours-28}) using the same number of components as FW. Similarly, \textbf{Ours-50} and \textbf{Ours-100} denote model with 50 and 100 components respectively. It is clear that our blendshape model largely outperforms FW model, regardless of the number of components to utilise.

Additionally, we plot a dozen of 3D expression transfer examples in Figure~\ref{fig:blendshapes}. Specifically, we calculated the facial deformation of each unseen expression and reconstruct using our blendshape and FW respectively. We then cast the reconstructed expression on a mean face~\cite{Booth2017} for visualisation. Note that we fixed the number of our expression components to be identical with FW. It is obvious that our blendshape model can faithfully reconstruct unseen expressive faces with correct expression meaning. Moreover, our recovered shapes contain more facial details, such as wrinkles between the eyebrows.
\subsection{Expression transfer}
Our expression blendshape model, together with the dynamic sequence, can be used to synthesize new motion sequences from one single 3D face, with or without texture. In particular, we reconstruct each expression of the dynamic sequence from the source actor using our blendshape model, and transfer these expressions to the target actor. We demonstrate this application via an exemplar video of Tom Cruise downloaded from Youtube. There are three processing steps involved: (1) we perform a 3D Morphable Model (3DMM) fitting~\cite{Booth20173DFM} on the given video, and select one 3D neutral face $\mathbf{T}$ among all the fitted meshes which will be our transfer target; (2) we select a 4D sequence $S=\{ \mathbf{S}_1, ..., \mathbf{S}_n \}$ that exhibits rich expressions from one subject in our database, compute the dense deformations with regard to the subject's neutral face and reconstruct all the expressions using our blendshape model; (3) finally we can apply the reconstructed expressions $\Delta s=\{ \Delta\mathbf{s}_1, ..., \Delta\mathbf{s}_n \}$ to the target $\mathbf{T}$ separately and obtain a new expressive sequence of Tom Cruise $T_{new}=\{ \mathbf{T}+\Delta\mathbf{s}_1, ..., \mathbf{T}+\Delta\mathbf{s}_n \}$. For every newly generated mesh, we use the same UV texture from $\mathbf{T}$. There are quite a few elaborated methods to generate more realistic results, interested reader may refer to~\cite{obama_lips,face2face_Thies_16}. For entertainment purposes, we rewind the target video to the same length of source sequence, and render every synthetic meshes to the extended video. We select two dynamic sequences and perform the expression transfer separately, Figure~\ref{fig:expression_transfer} shows the screenshot of each demo video. Note that our method can be applied to any given 3D faces, provided that it has the same mesh topology as our model.


\section{Conclusion}

We have presented 4DFAB, the first large scale 4D facial expression database that contains both posed and spontaneous expressions, and can be used for biometric application as well as facial expression analysis. We demonstrate the usefulness of the database in a series of recognition and verification experiments. We investigate, for the first time, the use of identity-free dense shape deformation from posed/spontaneous expression sequences in biometric applications. Promising results are obtained with basic features and standard classifier, thus we believe that dynamic facial behaviours could be further exploited for face recognition and verification. Last but not the least, we build a powerful expression blendshape model from this database, which outperforms the state-of-the-art blendshape model. The database will be made publicly available for research purposes. 

{\small
\bibliographystyle{ieee}
\bibliography{4dfab}
}

\end{document}